\documentclass[10pt, a4paper]{article}

\usepackage{lrec-coling2024} 

\usepackage{hyperref}
\usepackage{booktabs}
\usepackage{tabularx}
\usepackage{multirow}
\usepackage{graphicx}
\graphicspath{ {./images/} }

\title{LiMe: a Latin Corpus of Late Medieval Criminal Sentences}

\name{Alessandra Bassani$^{1}$, Beatrice Del Bo$^{2}$, Alfio Ferrara$^{3}$, Marta Mangini$^{2}$, \\ {\bf \large Sergio Picascia$^{3}$, Ambra Stefanello$^{4}$} \\}

\address{
$^{1}$Università degli Studi di Milano, Department of Italian and Supranational Public Law \\
$^{2}$Università degli Studi di Milano, Department of Historical Studies \\
$^{3}$Università degli Studi di Milano, Department of Computer Science \\ 
$^{4}$Università degli Studi di Firenze, Department of History, Archaeology, Geography, Fine and \\ Performing Arts \\
}

\abstract{
The Latin language has received attention from the computational linguistics research community, which has built, over the years, several valuable resources, ranging from detailed annotated corpora to sophisticated tools for linguistic analysis. With the recent advent of large language models, researchers have also started developing models capable of generating vector representations of Latin texts. The performances of such models remain behind the ones for modern languages, given the disparity in available data. In this paper, we present the LiMe dataset, a corpus of 325 documents extracted from a series of medieval manuscripts called \textit{Libri sententiarum potestatis Mediolani}, and thoroughly annotated by experts, in order to be employed for masked language model, as well as supervised natural language processing tasks.
\\ 
\newline 
 }

\begin{document}

\maketitleabstract

\section{Introduction}
\label{introduction}

The manuscripts called \textit{Libri sententiarum potestatis Mediolani}, preserved at the \textit{Archivio Storico Civico and Biblioteca Trivulziana in Milan, Cimeli, 146-152}, represent all that remains of the documentation recorded in the late medieval period at the court of justice of the city of Milan. The seven manuscripts of the series cover the activity of the court during the years 1385, 1390-1392, 1397-1398, 1398-1399, 1400-1401, 1427 and 1428-1429, respectively, resulting in the delivery of approximately 3,000 criminal sentences\footnote{Throughout the article, the term ``sentence'' will be used with its meaning of \textit{a punishment that a judge gives to someone who has committed a crime}.} discussed in the presence of the Milanese judges, pronounced by the \textit{podestà}\footnote{A chief magistrate of a medieval Italian town.} and publicly recorded by the notaries who worked at the court in the \textit{Loggia degli Osii}\footnote{A historical building of Milan, from whose balcony sentences and edicts were proclaimed by the Milanese judges.}. Although, as evident, the chronological span of each \textit{Liber} varies considerably according to the length of time each \textit{podestà} was in office, the structure, the material aspect and even the form employed in the drafting of these manuscripts present elements of a certain homogeneity and uniformity. This is due to the fact that the notaries in charge of assisting mayors and judges during trials recorded the sentences according to a pattern that is repeated almost unchanged in all manuscripts.

Each verdict, preceded by the verbal invocation - \textit{In nomine Domini, amen}\footnote{\textit{In the name of the Lord, amen.}} - is pronounced by the \textit{podestà} in accordance with the seigniorial decrees and statutes of the municipality of Milan. It contains the names of the accused, the narration of the legal proceeding, whether it was an \textit{inquisitio} or an accusation, with the salient phases of the trial and the final pronouncement. In addition to the sentences, whose pattern is formally identical for all defendants, there are also numerous subsequent interventions: e.g. annotations relating to receipts for full or partial payment of penalties or cancellations of sentences.

The \textit{Libri sententiarum potestatis Mediolani} are pivotal sources for law historians, like all Medieval and Early Modern trial outcomes preserved in the European archives: they allow us to measure the distance between the discipline established by \textit{statuta} and \textit{ius comune} and its actual application before the courts of medieval cities~\citep{padoa-schioppa-2017}. Indeed, the seven \textit{Libri} photograph the complex balance of social and political forces that characterised the city of Milan during the Visconti rule~\citep{gamberini2014companion}.

This documentary typology constitutes a source of great importance for historians of medieval law~\citep{storti-2021, valsecchi-2021, bassani-2021-4464998, isotton-2021-4465004, bianchi-riva-2021-4465006, minnucci-2021-4465070}, meanwhile fulfilling the same function for medievalists tout court. It provides inspiration for those who deal with political and institutional history, since it allows one to investigate in practice the dynamics of the exercise and management of power, the men, the methods and timing through which justice is administered, including through the selection of judges~\citep{pagnoni-2021-4464969}; at the same time, a collection of sentences issued by a city lord provides very useful elements for the study of society and economy, through the analysis and reconstruction of the type of crime, its scene and circumstances, the weapons used, the profiles of the people involved, including their reputation, qualification and profession.

In this article, we present the LiMe dataset, an annotated Latin corpus consisting of 325 judicial documents from the first volume of the \textit{Libri sententiarum potestatis Mediolani}. We illustrate the process undertaken for digitizing the documents and annotating them with detailed information, such as entities and relations, in order to make the manuscript more accessible and valuable to researchers. The paper is structured as follows: Section~\ref{motivation} provides the motivations behind this research; Section~\ref{relwork} outlines relative contributions in the field literature; in Section~\ref{dataset} we define how the data has been extracted and the final structure of the LiMe dataset; Section~\ref{applications} gives examples of possible statistical and machine learning applications; in Section~\ref{conclusion} we discuss the results and the future steps.

\section{Motivation}
\label{motivation}
The study of society through the filter of the judicial machine allows a better understanding of the objectives of ``political discipline'' and the effectiveness of this governing instruments~\citep{Campisi-2019, Campisi-2021}. At the same time, the registers of sentences still preserved in the archives of Italian cities of the last centuries of the Middle Ages, constitute a valuable field of research for those who deal with the history of gender in the medieval age~\citep{del-bo-2021-4464979, dean-2008}. The analysis of such documentation on the basis of the interpretative categories typical of this historiography benefits from the possibility of questioning the source on the characteristics of alleged victims and perpetrators, the type of condemnation/absolution, the granting of pardon (\textit{gratia}), the timing of the execution of the sentence, the type of crime, the weapons used, the place and circumstances of the offence (\textit{delictum}), single or group action, the presence of accomplices or leaders and their gender, the personal/familial condition, the words used to identify and define each person, to mention only a few aspects of the research. Starting from the identification modalities of women and men from the language of sentences, exploiting qualifying attributes, the source offers the possibility of dismantling stereotypes and historiographical clichés.

Despite their undoubted relevance, the \textit{Libri sententiarum potestatis Mediolani} have received little, if any, historiographical attention overall. In fact, they have not been taken into account in wide-ranging studies dedicated to the subject of the documentation issued by medieval Italian judicial bodies~\citep{giorgi2012documentazione, lett2021registri, dean-2007, vallerani2012medieval} and, until very recent years, few scholars have dealt with them specifically~\citep{verga1901sentenze, santoro1968offici, padoa1996giustizia, covini2012assenza}. The first manuscript in the series contains 126 criminal sentences pronounced by the \textit{podestà} of Milan Carlo Zen (1385). This manuscript was recently edited by~\citep{pizzi2021} and analysed in~\citep{bassani2021liber}. 

\section{Related Work}
\label{relwork}

Despite being a dead language with far less resources with respect to modern languages, Latin has recently received significant attention from the research community, in both the production of annotated datasets and the training of language-specific models. 

\subsection{Latin Corpora}
Several projects are currently dealing with the digitization and annotation of a considerable amount of Latin texts, often coming from different sources, with the purpose of being explored and exploited by history and linguistics scholars. Some of these corpora mainly present detailed syntactic and morphological annotations. It is the case of the five Latin Universal Dependencies\footnote{\url{https://universaldependencies.org/la/}} treebanks: PROIEL~\citep{proiel}, Perseus~\citep{perseus}, ITTB~\citep{ittb}, LLCT~\citep{llct}, UDante~\citep{udante}.
LatinISE~\citep{mcgillivray2013tools} is a Latin corpus for Sketch Engine, gathering documents from different websites; the corpus can be searched through the usage of tokens (13 million those present in the documents), or filtered on metadata, such as the author or the time period of each work. The LIRE~\citep{kavse2021classifying} dataset is another example of data integration, collecting Latin inscriptions dating back to the Roman Empire from two sources: the Epigraphic Database Heidelberg\footnote{\url{https://edh.ub.uni-heidelberg.de}} (EDH) and the Epigraphik Datenbank Clauss-Slaby\footnote{\url{http://www.manfredclauss.de}} (EDCS). The Opera Latina corpus~\citep{lasla}, created and maintained by the Laboratoire d’Analyse Statistique des Langues Anciennes (LASLA) includes 154 works from 19 classical Latin authors. The recent LiLa\footnote{\url{https://lila-erc.eu}}~\citep{lila} (Linking Latin) project has the object of building a common knowledge base, capable of describing several scattered Latin datasets with a unique vocabulary.

There are just a few cases of Latin corpora presenting detailed annotations for a specific task. The dataset presented in~\citep{Besnier-2021} contains proper nouns of people and places in three Medieval languages, Latin included; the dataset can be employed to build named entity recognition (NER) models for low-resource languages. Addressing the task of authorship analysis, MedLatinEpi and MedLatinLit~\citep{medlatin} are two datasets consisting of 294 and 30 curated texts, respectively, labelled with the respective author; MedLatinEpi texts are of epistolary nature, while MedLatinLit texts consist of literary comments and treatises about various subjects. 

Regarding legal texts, the Justinian's Digest has been digitized and included in a relational database~\citep{Ribary-2020}: the texts can be accessed and filtered, querying information about jurists, thematic sections and compositional structure.

\subsection{Latin Language Models}

In recent years, both non-contextual and contextual embedding models have been exploited for the representation of Latin text. In~\citep{burns-profiling} the authors train a word2vec model on a large Latin corpus, achieving state-of-art performances on synonym detection and inter-textual search. Latin BERT~\citep{bamman2020latin} is a contextual language model for Latin, trained on a large corpus spanning over twenty-two centuries; a fine-tuned version of Latin BERT~\citep{lendvai-wick-2022-finetuning} has been proposed for a word sense disambiguation task.

LatinCy~\citep{burns2023latincy} is an entire Latin NLP pipeline built for the Python library spaCy~\citep{spacy}: it consists of several models, capable of performing part-of-speech tagging, dependency parsing, and named entity recognition. Stanza~\citep{stanza} is a collection of tools and models for the linguistic analysis of many human languages, including Latin, trained on Universal Dipendencies treebanks. UDPipe~\citep{udpipe} is a pipeline for tokenization, tagging, lemmatization and dependency parsing, trainable on CoNLL-U files. 

Shared tasks are being proposed in order to foster research in the field of language technologies for Classical languages. The EvaLatin 2022 Evaluation Campaign~\citep{sprugnoli-etal-2022-overview} proposed three tasks relative to lemmatization, part-of-speech tagging, and features identification.

\section{Dataset}
\label{dataset}

LiMe\footnote{\url{https://doi.org/10.13130/RD_UNIMI/EN2TFH}}~\citep{lime} is a publicly available Latin corpus consisting not only of criminal sentences, but also of many additional notes gathered from the first manuscript of the \textit{Liber sententiarum potestatis Mediolani} (1385-1429), the oldest known registers of criminal sentences for the city of Milan. The original source, preserved in very good conditions and presenting just three mutilated texts, has been edited and transcribed in the curated edition~\citep{pizzi2021}. The texts have then been digitized and annotated in the context of the Fight Against Injustice Through Humanities (FAITH) project~\citep{ferrara2023faith}, whose main objective is to provide common tools and methodology for the collection, digitization and integration of different historical sources. For each document, named entities, relations between them and events have been manually identified; moreover, the texts have been classified depending on the type of document and, in case of criminal sentences, they have been segmented according to a predefined annotation schema. The result is a collection of 325 documents, made of 87110 tokens, in Latin language. The annotations, performed by a team of experts, have been organized according to a custom schema; an example of the annotations is provided in Section~\ref{annotation-structure}.

\subsection{Data Extraction}
The main source of information in the manuscript are the criminal sentences, gathered in dossiers and ordered according to an arbitrary number given from the curator, e.g. \textit{Sentenza I.1} refers to the first (1) judgment from the first (I) dossier. Each dossier is usually opened by a ``protocol'', i.e., a textual section in which the notary explicitly declares his identity and announces, following a very precise formulary, the name of the judge and \textit{podestà} who presided over the trials. The ``eschatocol'' is the section closing each dossier, where the notary refers to the group of judgments he has transcribed, citing the witnesses present. Additionally, there are three other types of sources, constituting supplementary information to the judgements: an ``addendum'' is a document added later to the text of the judgment, indicating further developments happened after the end of the trial; an ``insert'' is a piece of text, reported within a judgment or addendum, usually certifying orders received from the \textit{podestà}; finally, a ``news'' is an indirect evidence of an order or document that existed at the time but was not transcribed, useful in justifying decisions made by authority or actions taken by officials.

\begin{figure}[ht!]
    \centering
    \includegraphics[width=0.9\linewidth]{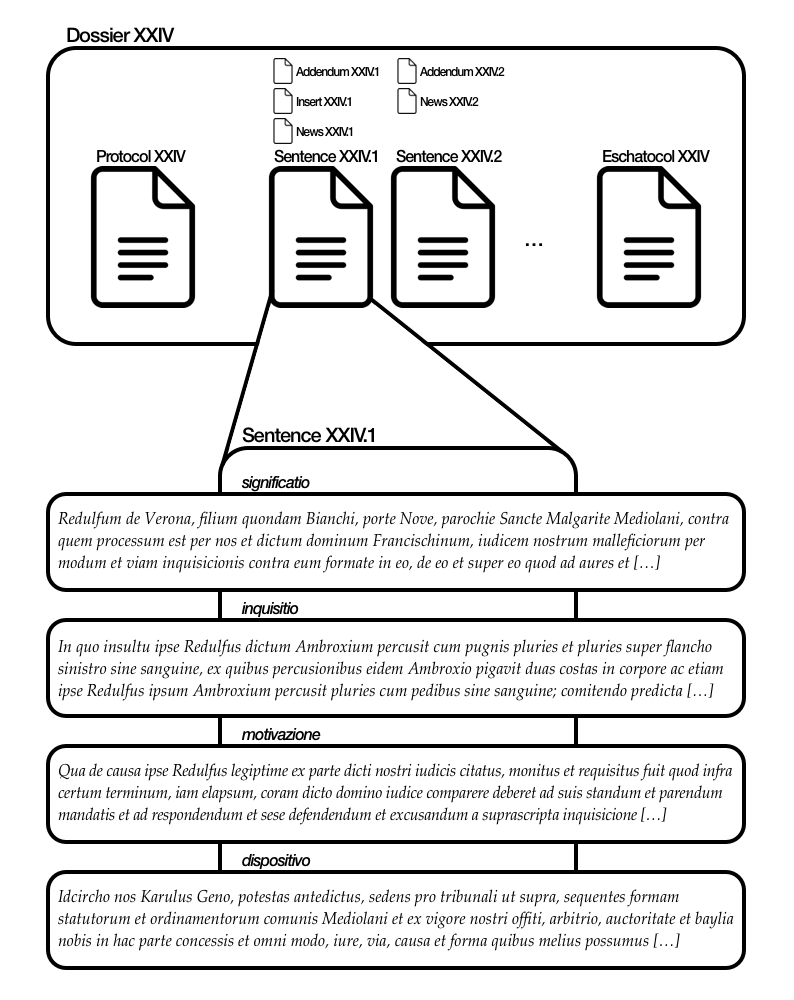}
    \caption{Example of the structure of a dossier and details on judgment's sections.}
    \label{fig:dossier-schema}
\end{figure}

The texts of criminal sentences, being them legal texts (thus with a rigid structure and a content pattern based on formulas), present the same sections and reflect a precise and largely stable structure. At the beginning, sometimes there it lies the \textit{significatio}, i.e., the communication of the misdemeanor(s) to the \textit{podestà} by a faithworthy person, the elder of the parish, in charge of the surveillance of a living area; this communication, however, did not always occur, so it is not always found in the text. The following part of the judgment, the \textit{inquisitio}, narrates the events that occurred as they were reconstructed: here, the details regarding each misdemeanor (\textit{misfatto}) are reported, such as the criminal offences, the perpetrator of the violence, the victim and any item involved. The motivational section (\textit{motivazioni}), usually introduced by the words \textit{qua de causa} (``the cause of''), \textit{et predicta} (``and the aforesaid'') or \textit{et constat nobis} (``and it is agreed with us''), states the reason why the verdict was reached. Finally, the last part of the sentence consists of the decision (\textit{dispositivo}) of conviction or acquittal and, in the former case, also of the type and amount of punishment; it generally begins with the word \textit{idcirco} (``therefore'', ``about that''). A summary of the structure of a typical dossier with details on the form of a judgment is depicted in Figure~\ref{fig:dossier-schema}.

The text of each source, strictly written in Latin language, has been thoroughly studied by experts, combining the findings extracted from the text with their domain knowledge in order to provide accurate and detailed annotations about people, places and items. For each person involved in the facts, demographic and social information have been identified: name and nicknames, biological gender, social class (\textit{dominus}), profession, place of origin or residency, possible relationships with relatives, and roles played in the events. For instance, we know that \textit{Laurentius de Roncho}, also referred to as \textit{Beleius} and son of \textit{Belollus}, was murdered in March 1385 by \textit{Iohanollus de Raude}, also known as \textit{Barachinus}.

Knowledge about places is important to understand where crimes were being committed and the geographical origin of the criminals: places inside the city regard the \textit{parochiae} (parishes) and \textit{portae} (gates), that were used to divide the territory of Milan; places outside the city are used for both towns under the jurisdiction of Milan, and for cities inside or outside of Italy; finally, generic places are used to indicate where a misdemeanor has taken place, e.g., a public street or a private house. The murderer of \textit{Laurentius de Roncho} took place in a public street near its residency, in \textit{Parochia Sancti Babile foris}, \textit{Porta Horientalis}.

Within the narrative of a criminal event, it is possible to read about items used within an assault or that had been stolen by pickpockets, along with the indication of the body parts struck or striking. Additionally, for stolen artifacts, it is also specified their value, expressed in the currency of the time. For example, \textit{Iohanollus de Raude} struck \textit{Laurentius de Roncho} dead in the occipital bone (\textit{in capite de retro}) with a tuck (\textit{stocho}), an ancient type of longsword.

\subsection{Annotation Structure}
\label{annotation-structure}
The annotation activity has been performed by a team of domain experts, that defined and mutually agreed on the custom guidelines followed throughout the entire process. The resulting dataset consists of a collection of 325 documents, of which most comprise the Latin text, the document type, named entities, events, relations, and text segmentation labels. 

The documents are classified according to the six document types identified at the beginning of the previous section; the counts of documents for each type is reported in Table~\ref{tab:doctype-count}.

\begin{table}[ht!]
\begin{tabularx}{\linewidth}{@{}Xr@{}}
\toprule
\textbf{Type}       & \textbf{Count} \\ \midrule
Sentences  & 127   \\
Addendum   & 71    \\
News       & 48    \\
Protocol   & 30    \\
Insert     & 26    \\
Eschatocol & 23    \\ \bottomrule
\end{tabularx}
\caption{The list of document types ordered by number of occurrences.}
\label{tab:doctype-count}
\end{table}

Objects under the ``news'' type, given the fact that they are orders or information from non transcribed documents, do not have any text; thus, knowledge about ``news'' can be indirectly acquired from the text of another object they refer to, usually an ``addendum''. However, this knowledge is still reported in the ``news'' object in order to keep it logically distinct from the others.

In each document, there are eight types of named entity recognised: ``PERSON'' (e.g. \textit{Laurentius de Roncho}), ``PLACE'' (\textit{Parochia Sancti Babile foris}), ``DATE'' (\textit{01/03/1385-31/03/1385}), ``ITEM'' (\textit{stocho}), ``ANIMAL'' (\textit{equum brunum}, brown horse), ``MEASURE'' (\textit{valoris}, value), ``UNITY OF MEASURE'' (\textit{librarum imperialum}, imperial pounds), ``QUANTITY'' (\textit{viginti quinque}, twenty-five). For some of them, further sub-types have been defined, such as ``GIVEN NAME'' and ``NICKNAME'' for ``PERSON'', or ``CITY'' and ``CHURCH'' for ``PLACE''. The counts of named entities types and subtypes is reported in Table~\ref{tab:ner-count}; since the same named entity can occur in multiple documents, the counts refer to the unique occurrences in the entire dataset.

\begin{table}[ht!]
\begin{tabularx}{\linewidth}{@{}XXr@{}}
\toprule
\textbf{Type}           & \textbf{Sub-Type}     & \textbf{Count} \\ \midrule
PERSON                  & GIVEN NAME   & 721            \\
                        & NICKNAME     & 30             \\
                        & NAME VARIANT & 7              \\ [.5cm] 
PLACE                   & CHURCH       & 75             \\
                        & GENERIC      & 37             \\
                        & CITY         & 18             \\ [.5cm] 
DATE                    & DAY          & 105            \\
                        & RANGE        & 42             \\ [.5cm] 
ITEM                    & GENERIC      & 45             \\
                        & BONE         & 25             \\ [.5cm] 
QUANTITY                & GENERIC      & 38             \\ [.5cm] 
UNIT OF  \newline MEASURE       & GENERIC    & 10     \\ [1cm] 
MEASURE                 & GENERIC      & 7              \\ [.5cm] 
ANIMAL                  & GENERIC      & 3              \\ \bottomrule
\end{tabularx}
\caption{The list of named entity types and subtypes ordered by the number of unique occurrences.}
\label{tab:ner-count}
\end{table}

Events are the most complex structure in the dataset; each of them is characterised of a type, usually of a subtype, and one or more arguments. There are 5 types of events: ``TRIAL STAGE'', ``TRIAL INTEGRATION'', ``ESCHATOCOL'', ``OFFENCES'', and ``DEATH''. A type of event may have one or multiple subtypes, for a total of 37 event subtypes: for example, an event of type ``OFFENCES'' may be, among others, of subtype ``INSULT'', ``MURDER'' or ``THEFT''. Depending on its type and subtype, an event has a different set of attributes, each of them having a role and an entity playing that role: in a ``THEFT'' event, we expect to have a time and place of the event, a victim, a thief, and the object or quantity of money stolen.

\begin{figure}[ht!]
    \centering
    \includegraphics[width=0.99\linewidth]{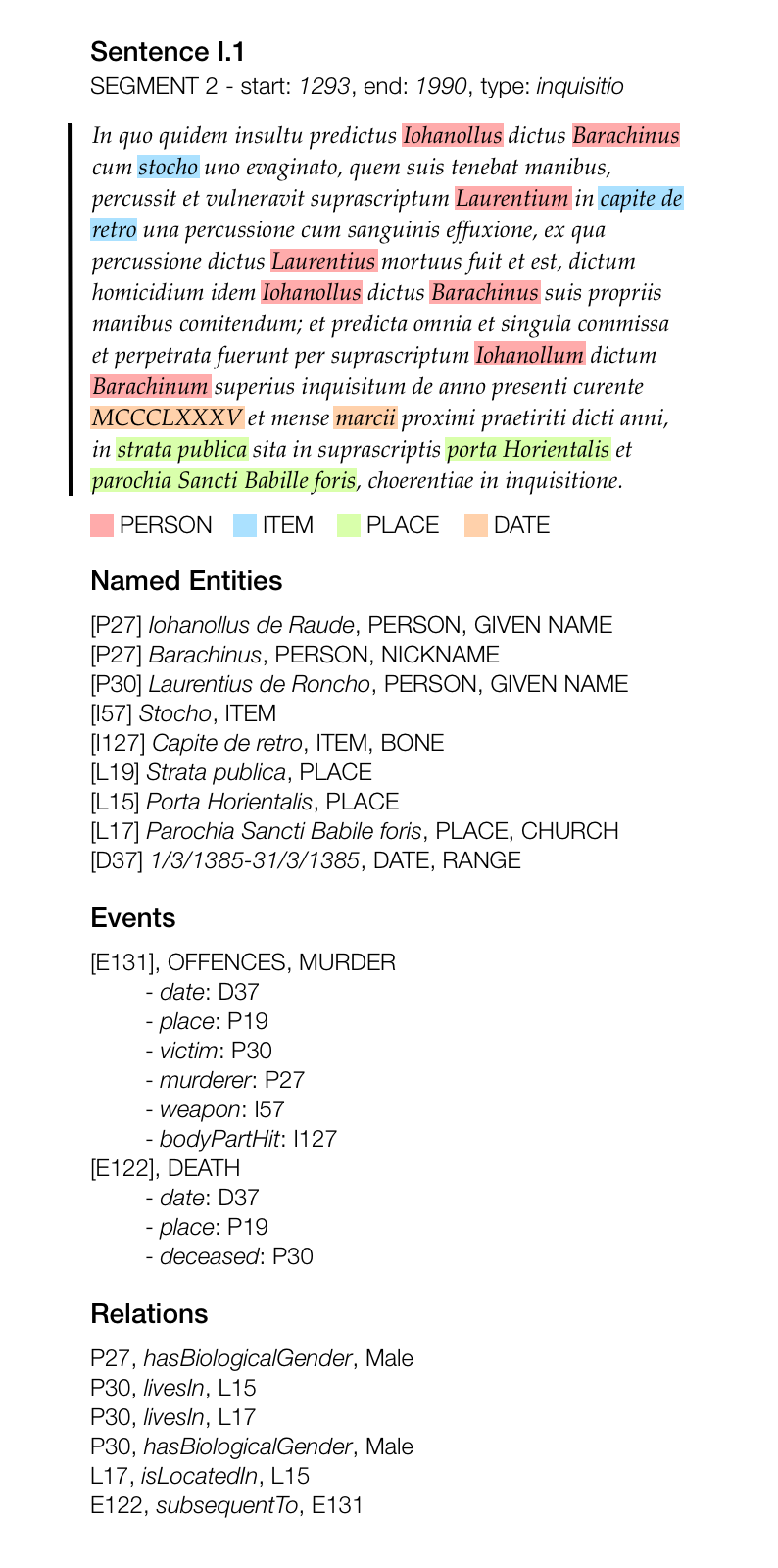}
    \caption{Example of the annotations of a segment taken from \textit{Sentence I.1}.}
    \label{fig:annotation}
\end{figure}

Relations between entities are defined by a triple of the form (``ENTITY1'', ``PREDICATE'', ``ENTITY2''), where ``ENTITY1'' is one of the named entities or events, ``PREDICATE'' defines the type of relation, and ``ENTITY2'' can be a named entity (or event) or a group. For instance, \textit{Laurentius de Roncho isSonOf Belollus} or \textit{Laurentius de Roncho hasBiologicalGender Male}. In the dataset there are 37 unique predicates, which define 3397 unique relations.

Finally, for documents of type ``sentences'', the text has been divided into segments, each of them classified with a label that specifies the section in which they appear, according to the annotation schema defined in the previous section: \textit{significatio}, \textit{inquisitio}, \textit{motivazioni}, \textit{dispositivo}. The segments are outlined by a starting and ending index, enclosing a specific span of text.

An example of all the annotations that can be found in a text is portrayed in Figure~\ref{fig:annotation}: this shows the amount of details that can be extracted even from a very short piece of text, like the one presented.

\section{Applications}
\label{applications}

In this section we provide examples of some possible use cases for the LiMe dataset, starting from simple exploratory analysis, that can be useful for medievalist researchers, to more elaborate Natural Language Processing (NLP) tasks.

\subsection{Exploratory Analysis}
\begin{figure}[ht!]
    \centering
    \includegraphics[width=0.9\linewidth]{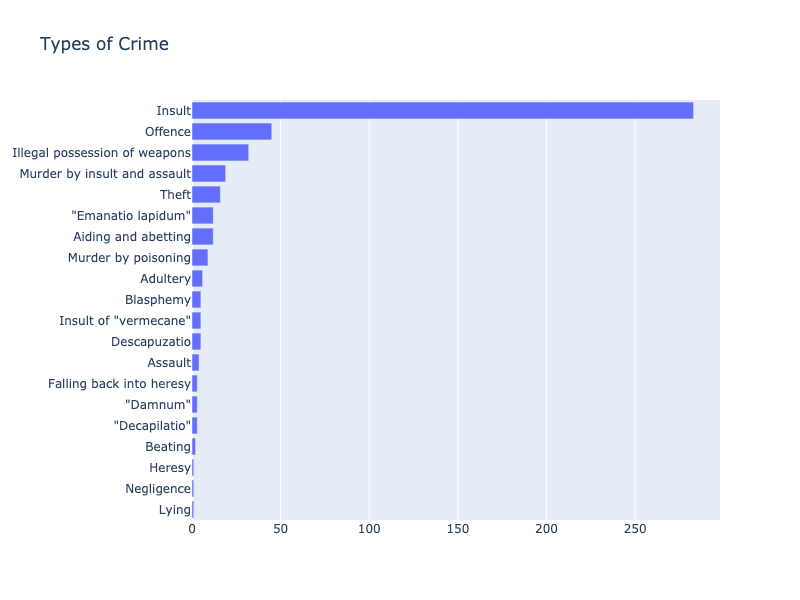}
    \caption{Number of criminal offences by type.}
    \label{fig:crime-types}
\end{figure}

The detail of annotation in the LiMe dataset allow for a methodological and technical study about social, demographic, judicial and economical aspects of the city of Milan in the XIV century. Extracting all the events of type ``OFFENCES'', and grouping them by subtype, it is possible to have an overview of the nature of crimes at the time. As shown in Figure~\ref{fig:crime-types}, beside some usual types of crime, such as insults, murders and thefts, there are some kinds of particular crimes, typical of that period, such as \textit{decapilatio}, the act of pulling someone's hair, and \textit{descapuzatio}, which consists in stealing a wool hat.

\begin{figure}[ht!]
    \centering
    \includegraphics[width=0.9\linewidth]{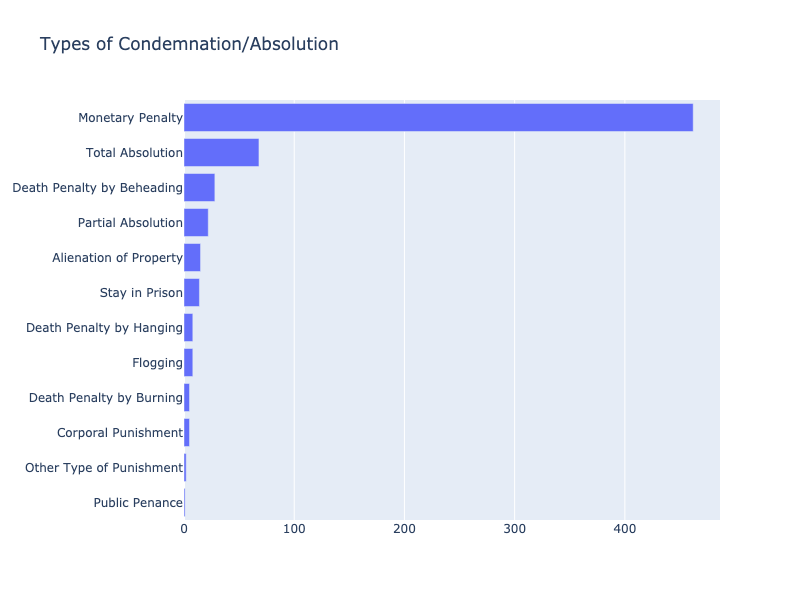}
    \caption{Number of condemnation/absolution by type.}
    \label{fig:cond-types}
\end{figure}

There are also some kinds of condemnation typical of the time, like flogging or corporal punishment (Figure~\ref{fig:cond-types}).

\begin{figure}[ht!]
    \centering
    \includegraphics[width=\linewidth]{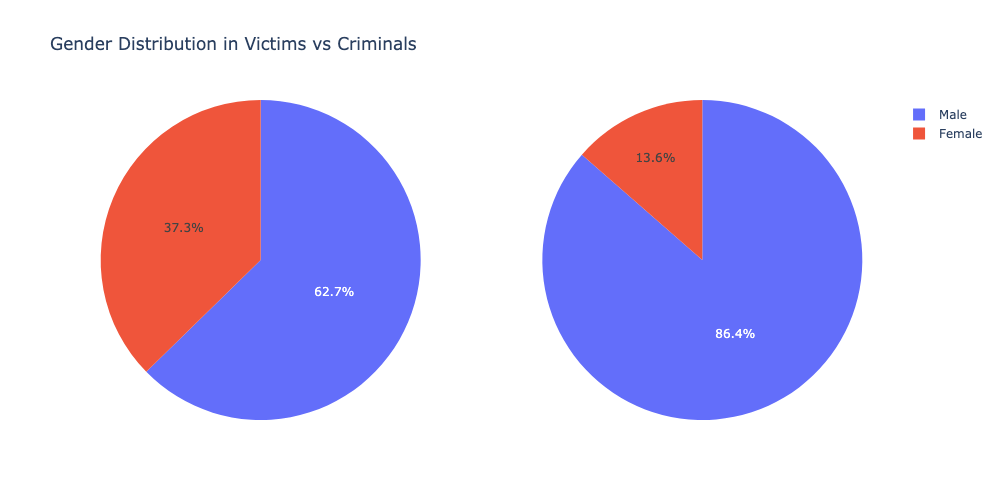}
    \caption{Distribution of males and females in victims (left) and criminals (right).}
    \label{fig:gender-distribution}
\end{figure}

It is also interesting to notice the difference in gender distribution of victims and criminals: despite them being mainly men in both cases, the percentage of females is almost triplicated when it comes to victims (Figure~\ref{fig:gender-distribution}).

\subsection{NLP Tasks}
Given the peculiarity of the dataset, we believe that LiMe can be employed for many machine learning tasks involving the usage of NLP techniques. Here we provide two examples of traditional problems: document classification and text segmentation.

\subsubsection{Document Classification}
\begin{figure}[ht!]
    \centering
    \includegraphics[width=0.9\linewidth]{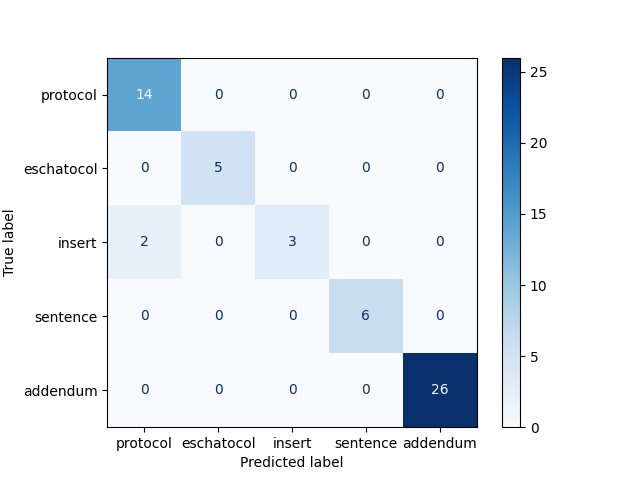}
    \caption{Confusion matrix for the document classification task.}
    \label{fig:doc-class}
\end{figure}

A document classification task regards the process of automatically assigning predefined labels to documents based on their content. For this reason, we decided to employ the 276 documents having a text, leaving out the ``news'' documents and ending up with five possible labels: ``addendum'', ``eschatocol'', ``insert'', ``protocol'', ``sentence''. We employ Latin BERT~\citep{bamman2020latin}, a contextual language model trained on a large corpus in Latin language, and fine-tune it on the training set (221 documents) for this specific classification task. The model achieves a weighted F1 score of 0.96 on the test set (55 documents), performing remarkably well on every class (Figure~\ref{fig:doc-class}).

\subsubsection{Text Segmentation}
\begin{figure}[ht!]
    \centering
    \includegraphics[width=0.9\linewidth]{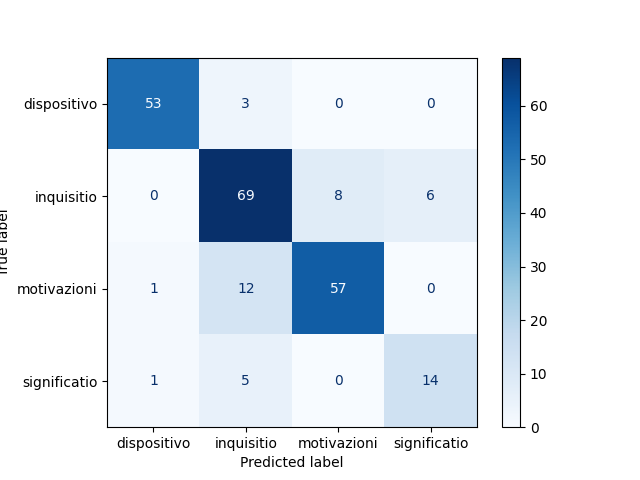}
    \caption{Confusion matrix for the text segmentation task.}
    \label{fig:text-seg}
\end{figure}

A text segmentation task consists in dividing a given text into meaningful and coherent segments based on an underline annotation schema. The documents interested by this task are the ``sentences'' that, together, are made of more than one thousand textual segments. Each of them has a section associated to it, according to the following schema: ``significatio'', ``inquisitio'', ``motivazioni'', ``dispositivo''. In order to solve the task, we employ Rewired Conditional Random Fields~\citep{ferrarapicasciariva2023}, a recent approach developed for the textual segmentation of Italian judgments, capable of working in a few-shot scenario, which is ideal given the low number of available observations. We train the above model on the segments of one hundred ``sentences'': the model achieves a weighted F1 score of 0.84 on the remaining 20\% of the dataset left out for evaluation purposes (Figure~\ref{fig:text-seg}).

\section{Conclusion}
\label{conclusion}

The \textit{Libri sententiarum potestatis Mediolani} are a valuable resource not only for scholars studying medieval law, but also for historians and linguists. The LiMe dataset proves how the digitisation and annotation of these kinds of sources allow for a methodological and technical analysis of the data, thanks to the usage of statistical and machine learning tools. In the future, we expect to: exploit the current dataset for more complex tasks, such as named entity recognition or event extraction; increase the number of annotated documents, with information coming from subsequent volumes of the \textit{Libri}, which are currently being examined by experts; extend the current annotations with features at syntactical and morphological levels.

\nocite{*}
\section{References}\label{sec:reference}

\bibliographystyle{lrec-coling2024-natbib}
\bibliography{bibliography}

\begin{thebibliography}{46}
\expandafter\ifx\csname natexlab\endcsname\relax\def\natexlab#1{#1}\fi

\bibitem[{{Bamman} and {Burns}(2020)}]{bamman2020latin}
David {Bamman} and Patrick~J. {Burns}. 2020.
\newblock \href {https://doi.org/10.48550/arXiv.2009.10053} {{Latin BERT: A
  Contextual Language Model for Classical Philology}}.
\newblock \emph{arXiv e-prints}, page arXiv:2009.10053.

\bibitem[{Bamman and Crane(2011)}]{perseus}
David Bamman and Gregory Crane. 2011.
\newblock The ancient greek and latin dependency treebanks.
\newblock In \emph{Language Technology for Cultural Heritage}, pages 79--98,
  Berlin, Heidelberg. Springer Berlin Heidelberg.

\bibitem[{Bassani(2021)}]{bassani-2021-4464998}
Alessandra Bassani. 2021.
\newblock \href {https://doi.org/10.5281/zenodo.4464998} {{Le assoluzioni nel
  Liber comunis potestatis Mediolani: riflessioni sull'ipotesi di una giustizia
  giusta}}.
\newblock \emph{Notariorum Itinera}, 7:177--204.

\bibitem[{Bassani et~al.(2021)Bassani, Calleri, and Mangini}]{bassani2021liber}
Alessandra Bassani, Marta Calleri, and Marta~L. Mangini. 2021.
\newblock Liber sententiarum potestatis mediolani (1385): Storia, diritto,
  diplomatica e quadri comparativi.
\newblock \emph{Notariorum Itinera}, 7.

\bibitem[{Bassani et~al.(2024)Bassani, Del~Bo, Ferrara, Mangini, Picascia, and
  Stefanello}]{lime}
Alessandra Bassani, Beatrice Del~Bo, Alfio Ferrara, Marta Mangini, Sergio
  Picascia, and Ambra Stefanello. 2024.
\newblock \href {https://doi.org/10.13130/RD{\_}UNIMI/EN2TFH} {{LiMe - Liber
  sententiarum potestatis Mediolani}}.

\bibitem[{Besnier and Mattingly(2021)}]{Besnier-2021}
Clément Besnier and William Mattingly. 2021.
\newblock \href {https://doi.org/10.5334/johd.36} {Named-entity dataset for
  medieval latin, middle high german and old norse}.
\newblock \emph{Journal of Open Humanities Data}, 7(0):23.

\bibitem[{Bianchi~Riva(2021)}]{bianchi-riva-2021-4465006}
Raffaella Bianchi~Riva. 2021.
\newblock \href {https://doi.org/10.5281/zenodo.4465006} {{Iniuria e insultus
  tra diritto e politica. Le offese alle magistrature comunali nella
  legislazione statutaria e nella prassi giudiziaria in età viscontea}}.
\newblock \emph{Notariorum Itinera}, 7:239--264.

\bibitem[{Burns(2023)}]{burns2023latincy}
Patrick~J. Burns. 2023.
\newblock \href {http://arxiv.org/abs/2305.04365} {Latincy: Synthetic trained
  pipelines for latin nlp}.

\bibitem[{Burns et~al.(2021)Burns, Brofos, Li, Chaudhuri, and
  Dexter}]{burns-profiling}
Patrick~J. Burns, James~A. Brofos, Kyle Li, Pramit Chaudhuri, and Joseph~P.
  Dexter. 2021.
\newblock \href {https://doi.org/10.18653/v1/2021.naacl-main.389} {Profiling of
  intertextuality in latin literature using word embeddings}.
\newblock \emph{Proceedings of the 2021 Conference of the North American
  Chapter of the Association for Computational Linguistics: Human Language
  Technologies}, pages 4900--4907.

\bibitem[{Campisi(2019)}]{Campisi-2019}
Luca Campisi. 2019.
\newblock \href {https://doi.org/10.54103/2611-318X/11539} {Prassi giudiziaria
  a vercelli nel xiv secolo}.
\newblock \emph{Studi di storia medioevale e di diplomatica - Nuova Serie},
  (2):131–150.

\bibitem[{Cecchini et~al.(2020)Cecchini, Korkiakangas, and Passarotti}]{llct}
Flavio~Massimiliano Cecchini, Timo Korkiakangas, and Marco Passarotti. 2020.
\newblock \href {https://aclanthology.org/2020.lrec-1.117} {A new {L}atin
  treebank for {U}niversal {D}ependencies: Charters between {A}ncient {L}atin
  and {R}omance languages}.
\newblock In \emph{Proceedings of the Twelfth Language Resources and Evaluation
  Conference}, pages 933--942, Marseille, France. European Language Resources
  Association.

\bibitem[{Corbara et~al.(2022)Corbara, Moreo, Sebastiani, and
  Tavoni}]{medlatin}
Silvia Corbara, Alejandro Moreo, Fabrizio Sebastiani, and Mirko Tavoni. 2022.
\newblock \href {https://doi.org/10.1145/3485822} {Medlatinepi and medlatinlit:
  Two datasets for the computational authorship analysis of medieval latin
  texts}.
\newblock \emph{Journal on Computing and Cultural Heritage}, 15(3):1--15.

\bibitem[{Covini(2012)}]{covini2012assenza}
Nadia Covini. 2012.
\newblock Assenza o abbondanza? la documentazione giudiziaria lombarda nei
  fondi notarili e nelle carte ducali (stato di milano, xiv-xv secolo).
\newblock \emph{La documentazione degli organi giudiziari}, pages 483--499.

\bibitem[{Dean(2007)}]{dean-2007}
Trevor Dean. 2007.
\newblock \href {https://doi.org/10.1017/CBO9780511496455} {\emph{Crime and
  Justice in Late Medieval Italy}}.
\newblock Cambridge University Press.

\bibitem[{Dean(2008)}]{dean-2008}
Trevor Dean. 2008.
\newblock \href
  {https://doi.org/https://doi.org/10.1111/j.1468-0424.2008.00526.x} {Theft and
  gender in late medieval bologna}.
\newblock \emph{Gender \& History}, 20(2):399--415.

\bibitem[{Del~Bo(2021)}]{del-bo-2021-4464979}
Beatrice Del~Bo. 2021.
\newblock \href {https://doi.org/10.5281/zenodo.4464979} {{Tutte le donne (del
  registro) del podestà fra cliché e novità}}.
\newblock \emph{Notariorum Itinera}, 7:83--106.

\bibitem[{Denooz(2007)}]{lasla}
Joseph Denooz. 2007.
\newblock Opera latina: le nouveau site internet du lasla.
\newblock \emph{Journal of Latin Linguistics}, 9(3):21--34.

\bibitem[{Ferrara et~al.(2023{\natexlab{a}})Ferrara, Picascia, and
  Riva}]{ferrarapicasciariva2023}
Alfio Ferrara, Sergio Picascia, and Davide Riva. 2023{\natexlab{a}}.
\newblock Few-shot legal text segmentation via rewiring conditional random
  fields: A preliminary study.
\newblock In \emph{Advances in Conceptual Modeling}, pages 141--150, Cham.
  Springer Nature Switzerland.

\bibitem[{Ferrara et~al.(2023{\natexlab{b}})Ferrara, Picascia, Rocchetti,
  Varese et~al.}]{ferrara2023faith}
Alfio Ferrara, Sergio Picascia, Elisabetta Rocchetti, Gaia Varese, et~al.
  2023{\natexlab{b}}.
\newblock The {FAITH} project: integrated tools and methodologies for digital
  humanities.
\newblock \emph{Proceedings of the Statistics and Data Science Conference},
  pages 323--327.

\bibitem[{Flavio et~al.(2020)Flavio, Sprugnoli, Giovanni, Marco
  et~al.}]{udante}
Cecchini Flavio, Rachele Sprugnoli, Moretti Giovanni, Passarotti Marco, et~al.
  2020.
\newblock Udante: First steps towards the universal dependencies treebank of
  dante’s latin works.
\newblock In \emph{Proceedings of the Seventh Italian Conference on
  Computational Linguistics}, pages 99--105. Accademia University Press.

\bibitem[{Gamberini(2014)}]{gamberini2014companion}
Andrea Gamberini. 2014.
\newblock \emph{A Companion to Late Medieval and Early Modern Milan: The
  Distinctive Features of an Italian State}, volume~7.
\newblock Brill.

\bibitem[{Giorgi et~al.(2012)Giorgi, Moscadelli, and
  Zarrilli}]{giorgi2012documentazione}
Andrea Giorgi, Stefano Moscadelli, and Carla Zarrilli. 2012.
\newblock \href {https://books.google.it/books?id=xazZnQEACAAJ} {\emph{La
  documentazione degli organi giudiziari nell'Italia tardo-medievale e moderna:
  atti del convegno di studi, Siena, Archivio di Stato, 15-17 settembre 2008}}.
\newblock Number v. 1 in Pubblicazioni degli Archivi di Stato. Saggi.
  Ministerio per i beni e le attivit{\`a} culturali, direzione generale per gli
  archivi.

\bibitem[{Haug and J{\o}hndal(2008)}]{proiel}
Dag~TT Haug and Marius J{\o}hndal. 2008.
\newblock Creating a parallel treebank of the old indo-european bible
  translations.
\newblock In \emph{Proceedings of the second workshop on language technology
  for cultural heritage data (LaTeCH 2008)}, pages 27--34.

\bibitem[{Honnibal et~al.(2020)Honnibal, Montani, Van~Landeghem, and
  Boyd}]{spacy}
Matthew Honnibal, Ines Montani, Sofie Van~Landeghem, and Adriane Boyd. 2020.
\newblock \href {https://doi.org/10.5281/zenodo.1212303} {spacy:
  Industrial-strength natural language processing in python}.

\bibitem[{Isotton(2021)}]{isotton-2021-4465004}
Roberto Isotton. 2021.
\newblock \href {https://doi.org/10.5281/zenodo.4465004} {{La repressione dei
  reati di furto e rapina nel Liber sententiarum potestatis Mediolani del 1385:
  acquisizioni e questioni aperte}}.
\newblock \emph{Notariorum Itinera}, 7:205--238.

\bibitem[{Ka{\v{s}}e et~al.(2021)Ka{\v{s}}e, He{\v{r}}m{\'a}nkov{\'a}, and
  Sobotkov{\'a}}]{kavse2021classifying}
Vojtěch Ka{\v{s}}e, Petra He{\v{r}}m{\'a}nkov{\'a}, and Adéla Sobotkov{\'a}.
  2021.
\newblock Classifying latin inscriptions of the roman empire: A
  machine-learning approach.
\newblock In \emph{Proceedings of the Conference on Computational Humanities
  Research 2021CEUR-WS}, volume 2989, pages 123--135.

\bibitem[{Lendvai and Wick(2022)}]{lendvai-wick-2022-finetuning}
Piroska Lendvai and Claudia Wick. 2022.
\newblock \href {https://aclanthology.org/2022.cogalex-1.5} {Finetuning {L}atin
  {BERT} for word sense disambiguation on the thesaurus linguae latinae}.
\newblock In \emph{Proceedings of the Workshop on Cognitive Aspects of the
  Lexicon}, pages 37--41, Taipei, Taiwan. Association for Computational
  Linguistics.

\bibitem[{Lett(2021)}]{lett2021registri}
Didier Lett. 2021.
\newblock \href {https://books.google.it/books?id=ddQxzgEACAAJ} {\emph{I
  Registri Della Giustizia Penale Nell'Italia Dei Secoli XII-XV}}.
\newblock Ecole fran{\c{c}}aise de Rome.

\bibitem[{Luca(2021)}]{Campisi-2021}
Campisi Luca. 2021.
\newblock \emph{L’impatto sociale. I protagonisti delle pratiche giudiziarie
  a Vercelli fra XIV e XV secolo}.
\newblock Phd thesis, Università degli Studi di Milano.

\bibitem[{McGillivray and Kilgarriff(2013)}]{mcgillivray2013tools}
Barbara McGillivray and Adam Kilgarriff. 2013.
\newblock Tools for historical corpus research, and a corpus of latin.
\newblock \emph{New methods in historical corpus linguistics}, 1(3):247--257.

\bibitem[{Minnucci(2021)}]{minnucci-2021-4465070}
Giovanni Minnucci. 2021.
\newblock \href {https://doi.org/10.5281/zenodo.4465070} {{Intorno al Liber
  sententiarum potestatis Mediolani e ad altre fonti giudiziarie. Alcune note
  conclusive}}.
\newblock \emph{Notariorum Itinera}, 7:373--380.

\bibitem[{Padoa-Schioppa(1996)}]{padoa1996giustizia}
Antonio Padoa-Schioppa. 1996.
\newblock \emph{La giustizia milanese nella prima et{\`a} viscontea
  (1277-1300)}.
\newblock Giuffr{\`e}.

\bibitem[{Padoa-Schioppa(2017)}]{padoa-schioppa-2017}
Antonio Padoa-Schioppa. 2017.
\newblock \href {https://doi.org/10.1017/9781316848227} {\emph{A History of Law
  in Europe: From the Early Middle Ages to the Twentieth Century}}.
\newblock Cambridge University Press.

\bibitem[{Pagnoni(2021)}]{pagnoni-2021-4464969}
Fabrizio Pagnoni. 2021.
\newblock \href {https://doi.org/10.5281/zenodo.4464969} {{Selezione e
  circolazione dei giudici ai malefici nel dominio visconteo fra Tre e
  Quattrocento}}.
\newblock \emph{Notariorum Itinera}, 7:61--81.

\bibitem[{Passarotti(2019)}]{ittb}
Marco Passarotti. 2019.
\newblock \href {https://doi.org/doi:10.1515/9783110599572-017} {\emph{The
  Project of the Index Thomisticus Treebank}}, pages 299--320. De Gruyter Saur,
  Berlin, Boston.

\bibitem[{Passarotti et~al.(2020)Passarotti, Mambrini, Franzini, Cecchini,
  Litta, Moretti, Ruffolo, and Sprugnoli}]{lila}
Marco Passarotti, Francesco Mambrini, Greta Franzini, Flavio~Massimiliano
  Cecchini, Eleonora Litta, Giovanni Moretti, Paolo Ruffolo, and Rachele
  Sprugnoli. 2020.
\newblock Interlinking through lemmas. the lexical collection of the lila
  knowledge base of linguistic resources for latin.
\newblock \emph{Studi e Saggi Linguistici}, 58(1):177--212.

\bibitem[{Pizzi(2021)}]{pizzi2021}
Pier~Francesco Pizzi. 2021.
\newblock \emph{Liber sententiarum potestatis Mediolani (1385), Edizione
  critica}.
\newblock Società Ligure di Storia Patria.

\bibitem[{Qi et~al.(2020)Qi, Zhang, Zhang, Bolton, and Manning}]{stanza}
Peng Qi, Yuhao Zhang, Yuhui Zhang, Jason Bolton, and Christopher~D. Manning.
  2020.
\newblock Stanza: A {Python} natural language processing toolkit for many human
  languages.
\newblock In \emph{Proceedings of the 58th Annual Meeting of the Association
  for Computational Linguistics: System Demonstrations}.

\bibitem[{Ribary(2020)}]{Ribary-2020}
Marton Ribary. 2020.
\newblock \href {https://doi.org/10.5334/johd.17} {A relational database of
  roman law based on justinian’s digest}.
\newblock \emph{Journal of Open Humanities Data}, 6(1):5.

\bibitem[{Santoro(1968)}]{santoro1968offici}
C.~Santoro. 1968.
\newblock \href {https://books.google.it/books?id=5HYKAQAAIAAJ} {\emph{Gli
  offici del comune di Milano e del dominio visconteo sforzesco (1216-1515).}}
\newblock 1. collana: Monografie, ricerche ausilierie, opere strumentali. A.
  Guiffr{\`e}.

\bibitem[{Sprugnoli et~al.(2022)Sprugnoli, Passarotti, Cecchini, Fantoli, and
  Moretti}]{sprugnoli-etal-2022-overview}
Rachele Sprugnoli, Marco Passarotti, Flavio~Massimiliano Cecchini, Margherita
  Fantoli, and Giovanni Moretti. 2022.
\newblock \href {https://aclanthology.org/2022.lt4hala-1.29} {Overview of the
  {E}va{L}atin 2022 evaluation campaign}.
\newblock In \emph{Proceedings of the Second Workshop on Language Technologies
  for Historical and Ancient Languages}, pages 183--188, Marseille, France.
  European Language Resources Association.

\bibitem[{Storti(2021)}]{storti-2021}
Claudia Storti. 2021.
\newblock \href {https://doi.org/10.5281/zenodo.4464938} {1385: un anno tra
  politica e giustizia a milano}.
\newblock \emph{Notariorum Itinera}, 7:7--31.

\bibitem[{Straka(2018)}]{udpipe}
Milan Straka. 2018.
\newblock \href {https://doi.org/10.18653/v1/K18-2020} {{UDP}ipe 2.0 prototype
  at {C}o{NLL} 2018 {UD} shared task}.
\newblock In \emph{Proceedings of the {C}o{NLL} 2018 Shared Task: Multilingual
  Parsing from Raw Text to Universal Dependencies}, pages 197--207, Brussels,
  Belgium. Association for Computational Linguistics.

\bibitem[{Vallerani(2012)}]{vallerani2012medieval}
M.~Vallerani. 2012.
\newblock \href {https://books.google.it/books?id=WxwLh97zUpMC} {\emph{Medieval
  Public Justice}}.
\newblock Studies in Medieval \& Early Mo. Catholic University of America
  Press.

\bibitem[{Valsecchi(2021)}]{valsecchi-2021}
Chiara Valsecchi. 2021.
\newblock \href {https://doi.org/10.5281/zenodo.4464996} {{«Per viam
  inquisicionis». Note sul processo criminale a Milano in un'età di
  transizione}}.
\newblock \emph{Notariorum Itinera}, 7:127--176.

\bibitem[{Verga(1901)}]{verga1901sentenze}
E.~Verga. 1901.
\newblock \href {https://books.google.it/books?id=AdMnAAAAYAAJ} {\emph{Le
  sentenze criminali dei podest{\`a} milanesi 1385-1429: appunti per la storia
  della giustizia punitiva in Milano}}.
\newblock P. Confalonieri.

\end{thebibliography}

\end{document}